\documentclass[twocolumn,letterpaper]{IEEEAerospaceCLS}  

\usepackage[]{graphicx}    
\usepackage[colorlinks=true, urlcolor=blue, linkcolor=black, citecolor=black]{hyperref} 
\newcommand{\ignore}[1]{}  
\usepackage{tikz}
\usepackage{amsmath}
\usepackage{amssymb}
\usetikzlibrary{arrows.meta, positioning}
\begin{document}
\title{LOGIC: An LLM Benchmark for Intent-Grounded Change Impact in Aerospace Electrical Systems}

\author{%
\begin{tabular}{@{}c@{\hspace{2em}}c@{}}
Muhammad Faraz Shoaib$^{*}$ & Muhammad Qasim$^{*}$\\
School of Computing & School of Computing\\
Wichita State University & Wichita State University\\
1845 Fairmount St. & 1845 Fairmount St.\\
Wichita, KS 67260 & Wichita, KS 67260\\
mxshoaib3@shockers.wichita.edu & mxqasim@shockers.wichita.edu\\[2ex]
Raisulhaq Mohammed Rizwan & Rahmatullah Safdar\\
School of Computing & Department of Biomedical Engineering\\
Wichita State University & Wichita State University\\
1845 Fairmount St. & 1845 Fairmount St.\\
Wichita, KS 67260 & Wichita, KS 67260\\
fxraisulhaqmohammedrizwan@shockers.wichita.edu & rusafdar@shockers.wichita.edu\\[2ex]
Abdul Aleem Mohammed & Muzammil Adnan Shaik\\
School of Computing & Department of Industrial Engineering\\
Wichita State University & Wichita State University\\
1845 Fairmount St. & 1845 Fairmount St.\\
Wichita, KS 67260 & Wichita, KS 67260\\
axmohammed44@shockers.wichita.edu & mxshaik20@shockers.wichita.edu
\end{tabular}
\thanks{\footnotesize $^{*}$These authors contributed equally to this work.}
\thanks{\footnotesize 979-8-3195-1048-8/27/$\$31.00$ \copyright2027 IEEE}
}

\maketitle

\thispagestyle{plain}
\pagestyle{plain}

\begin{abstract}
Aerospace electrical-design revisions can contain multiple genuine changes, although an engineering request may authorize only a subset. Propagating every detected difference can therefore produce overly broad impact reports. We present LOGIC, a controlled benchmark and evaluation framework in which locally deployable language models ground a request in a deterministic candidate-change inventory before selected changes are propagated through a typed electrical traceability graph. This separation permits candidate-selection errors to be distinguished from downstream propagation errors. LOGIC contains 168 scenarios, including 144 selection and 24 abstention cases. We evaluate three 7--8B models against intent-agnostic, lexical, and structured-evidence methods, with an oracle-root upper bound. On 96 explicitly anchored selection cases, gate-only structured evidence achieves candidate F1 of 1.0000, compared with 0.9677 for token-lexical matching. On 12 relational-paraphrase cases, token-lexical F1 is 0.1772 and gate-only F1 is 0.0000, compared with 0.5000--0.6400 for the large language models. Model grounding degrades as candidate inventories grow from 4 to 64 changes, while affected-element and typed-path accuracy remain comparatively stable when frozen selections are replayed over graphs of approximately 1K to 100K nodes. Strict evidence gating suppresses false positives but can remove correct semantic selections. An exploratory evidence-empty abstention policy raises strict abstention accuracy to 0.6667 for all three models and reduces unsafe-report rates to 0.1667, while decreasing answerable-case coverage by 16.0--27.1 percentage points. Four of six conflicting requests remain unsafe for each model. Evaluation uses 504 frozen model predictions scored offline against separately stored reference labels. These findings support combining literal evidence and language-model reasoning with engineering review when intent cannot be established reliably.
\end{abstract}

\tableofcontents

\section{Introduction}
\label{sec:introduction}

Aerospace electrical designs evolve through equipment substitutions, wiring updates, interface changes, and verification-driven corrections. A single revision may therefore contain several genuine changes made for different purposes. Change-impact analysis traces an initiating modification to connected components, interfaces, requirements, verification artifacts, and documents~\cite{Clarkson2004,Brahma2023,Mordaschew2023}. In many such methods, the initiating change is already known. When an engineering request refers to only part of a revision, that initiating change must first be identified.

Consider a requested connector-pin reassignment for a navigation-related cooling load made alongside an unrelated wire-gauge update and equipment revision. All three are genuine revision differences, but propagating all of them would produce unrelated impacts. An LLM may interpret the request yet still select a similar distractor. The central task is to ground the request in the correct subset of observed changes before graph propagation.

We present LOGIC, a benchmark and evaluation framework for this task. Deterministic differencing of revisions before and after produces a candidate inventory C. Given request r, a grounding method selects candidate set $\widehat{S}\subseteq\mathcal{C}$ or abstains. Selected candidates are resolved to roots and propagated deterministically through a typed electrical traceability graph G. This separation makes it possible to examine errors in intent selection and downstream propagation independently.

LOGIC contains 168 controlled scenarios: 144 answerable requests and 24 requests requiring abstention. The cases include literal identifiers, functional and relational descriptions, multi-change requests, concurrent distractors, and ambiguous or conflicting intent. We evaluate Qwen2.5--7B, Llama3.1--8B, and Mistral--7B against lexical, structured-evidence, and hybrid grounding methods. The evaluation examines grounding accuracy, the effect of candidate load, evidence gating, abstention, and propagation accuracy across graph sizes.

We ask how accurately the methods ground requests (RQ1), how candidate load affects grounding (RQ2), when deterministic evidence helps or suppresses LLM selections (RQ3), whether evidence-informed abstention reduces unsafe reporting (RQ4), and how propagation accuracy changes with graph size (RQ5).

The results show complementary capabilities. Deterministic matching is strong when requests provide distinctive literal evidence, while the evaluated LLMs identify some relational intents that lexical matching misses. Evidence gating suppresses false positives but can remove correct semantic selections; abstention remains difficult, particularly when evidence conflicts.

The principal contributions are:
\begin{enumerate}
    \item An intent-grounded formulation that separates request-conditioned candidate selection from deterministic typed propagation.
    \item LOGIC, a controlled 168-scenario LLM benchmark with concurrent revision changes, semantic requests, and safety-oriented abstention cases.
    \item An oracle-separated evaluation of three local LLMs and deterministic and hybrid baselines, including candidate-load, graph-size, and abstention analyses.
\end{enumerate}

\section{Background and Related Work}

\label{sec:related-work}
 
Research relevant to LOGIC spans engineering change-impact analysis, typed traceability, natural-language grounding, and selective LLM decision making. These areas provide methods for identifying affected elements, but they differ in what is supplied as the initiating change. LOGIC focuses on an engineering request that must first be grounded in one or more genuine changes observed between two design revisions.
 
\subsection{Engineering Change Impact and Traceability}
 
Engineering change-impact analysis estimates how a modification may affect other system elements. Clarkson \textit{et al.} model the likelihood and consequence of change propagation between components~\cite{Clarkson2004}. Reviews by Brahma and Wynn and by Mordaschew \textit{et al.} describe a range of representations and techniques for analyzing propagation in complex products~\cite{Brahma2023,Mordaschew2023}. Such methods generally begin with a change or initiating element that has already been identified. LOGIC examines the preceding decision: which of several changes present in a revision are supported by a particular request?
 
Traceability methods provide the structure needed once an initiating change is known. Nejati \textit{et al.} use model-based relationships for design slicing and safety-related traceability; their subsequent work combines slicing with language-based ranking to reduce the elements requiring review~\cite{Nejati2012,Nejati2016}. Electrical and aircraft representations likewise provide relevant structures: VEC specifies electrical-system data exchange, CPACS represents aircraft configuration data, and WireViz documents individual cables, harnesses, and connector pinouts~\cite{VEC,CPACS,WireViz}. These representations inform the types of entities and relations that can be traced, but do not themselves evaluate request-conditioned selection among changes in an observed revision.
 
\subsection{Natural-Language Grounding and LLM-Based Impact Analysis}
 
Natural-language methods have been used to identify requirements affected by a change and, more recently, to estimate the impact of a supplied requirements change with LLMs~\cite{Arora2015,Etezadi2026}. ContCRIA is a particularly close staged approach: it interprets a change request against a structured requirements specification and uses model relationships to produce an impact report~\cite{Rajbhoj2025}. Its input is a request and a specification model. LOGIC instead presents the request alongside multiple changes already observed between two revisions and evaluates whether the intended candidate set is selected or the method abstains.
 
Ripple also uses change intent in impact analysis. It combines a natural-language description with a supplied seed edit location, then expands and refines the predicted impact set in a software repository~\cite{Yadavally2026}. LOGIC does not supply the initiating graph root: a grounding method must select it from competing revision candidates before deterministic graph propagation. We use \emph{intent-grounded} to denote this setting, in which roots are selected rather than supplied. Von Hei{\ss}en \textit{et al.} use a natural-language change request to retrieve SysML~v2 model content before an LLM proposes validated model edits~\cite{vonHeissen2026}; that system is authoring-oriented, whereas LOGIC is read-only. Tikhonenko \textit{et al.} give LLMs a known engineering change and ask for affected components end to end~\cite{Tikhonenko2026}; because selection and propagation occur in one response, errors cannot be attributed to either stage. LOGIC's controlled design isolates request-to-candidate grounding, abstention, and downstream typed propagation so that errors at these stages can be measured separately.
 
\subsection{Observed-Change Linking and Untangling}
 
Software-engineering studies offer related ways to connect natural-language intent with observed changes. Issue--commit linking matches an issue to relevant commits; plausible unrelated candidates make this task more difficult~\cite{Huang2025EasyLink}. Commit untangling instead separates a bundled diff into distinct concerns using lexical, structural, interactive, or model-based signals~\cite{Herzig2013,Partachi2020Flexeme,Shen2021SmartCommit,Li2022UTANGO,Xu2025Composite,Hou2026ColaUntangle}. These tasks clarify the challenge posed by concurrent changes. LOGIC asks a different question: given an external engineering request and a set of detected electrical-design changes, which subset should initiate impact propagation? Its output can also be \texttt{abstain} when the request cannot be grounded.
 
\subsection{Hybrid Decision Making and Abstention}
 
Hybrid systems assign different roles to language models and structured tools. DUCTILE, for example, combines LLM-based coordination with engineering tools in an aerospace analysis setting~\cite{PradasGomez2026}. LOGIC uses an LLM or baseline to select candidate changes, while graph-root resolution and typed propagation remain deterministic. This separation limits where model-generated uncertainty enters the impact report.
 
Declining to make an unsupported prediction is a longstanding concern in classification and selective prediction~\cite{Chow1970,ElYaniv2010,Geifman2017}. Work on LLM abstention and agreement-based uncertainty provides further motivation for testing when a model should withhold an answer~\cite{Wen2025Abstention,Feng2024Abstain}. In LOGIC, abstention is an explicit request-level outcome: selecting an unsupported change can initiate an incorrect downstream report, while excessive abstention prevents analysis of answerable requests. The evaluation therefore measures both unsafe reporting and the coverage cost of abstention policies.
 
\subsection{Positioning of LOGIC}
 
Prior methods may propagate a known initiating change, ground a request against a specification model, link a request to whole software commits, or partition a diff into concerns. LOGIC evaluates request-conditioned selection from multiple genuine changes observed between Bbefore and Bafter. A method selects $\widehat{S}\subseteq\mathcal{C}$ or abstains; selected candidates are then resolved to roots and propagated deterministically through typed graph G. This formulation permits the LLM's intent-grounding behavior to be evaluated separately from downstream propagation. To our knowledge, no prior work evaluates request-conditioned selection with abstention over observed changes in a typed aerospace-electrical traceability graph.

\section{Intent-Grounded Change-Impact Method}
\label{sec:method}
 
\begin{figure*}[t]
\centering
\includegraphics[width=\textwidth]{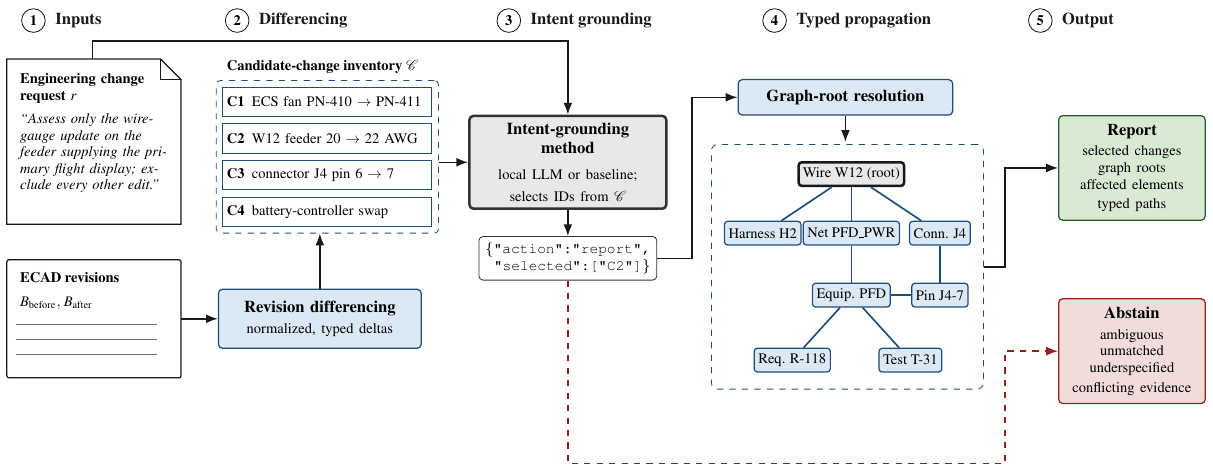}
\caption{LOGIC pipeline. Deterministic revision differencing
produces a candidate-change inventory from two ECAD
revisions. The grounding method selects request-relevant
candidates or abstains. Selected candidates are resolved to
roots and propagated through a typed electrical traceability
graph; abstention bypasses propagation.}
\label{logic_pipeline}
\end{figure*}

LOGIC separates change-impact analysis into two stages:
intent grounding and deterministic impact propagation, as
shown in Fig.~\ref{logic_pipeline}. An intent-grounding
method receives an engineering request and an inventory of
changes observed between two ECAD revisions. It selects
request-supported changes or abstains. Selected changes are
then resolved to roots in a typed electrical traceability
graph and propagated through permitted relationships.
 
\subsection{Problem Formulation}
\label{subsec:problem-formulation}
 
Let $r$ denote a natural-language engineering change
request. Let $B_{\mathrm{before}}$ and
$B_{\mathrm{after}}$ denote the normalized ECAD design
states before and after a revision. Deterministic
differencing produces the candidate inventory
\begin{equation}
\mathcal C=\Delta(B_{\mathrm{before}},B_{\mathrm{after}})
=\mathcal C_{\mathrm{add}}\cup
\mathcal C_{\mathrm{remove}}\cup
\mathcal C_{\mathrm{modify}},
\label{eq:candidate-extraction}
\end{equation}
where the three sets contain additions, removals, and
modifications, respectively. For identity-matched entities,
a modified entity may yield more than one atomic candidate.
Writing $\mathcal X$ for the set of matched entity pairs whose
attributes differ,
\begin{equation}
\mathcal C_{\mathrm{modify}}
=\bigcup_{(x,y)\in\mathcal X}\delta(x,y),
\label{eq:modified-candidates}
\end{equation}
where $\delta(x,y)$ is the \emph{set} of normalized
candidate records generated from the differences between
$x$ and $y$. Thus $\mathcal C=\{c_1,\ldots,c_n\}$ contains
the observed revision changes. For engineering request
$r$, the intended reference set is $S^*\subseteq\mathcal C$.
 
An intent-grounding method receives $(r,\mathcal C)$ and
returns an action and a predicted candidate set:
\begin{equation}
f(r,\mathcal C)\rightarrow(\widehat a,\widehat S),
\qquad
\widehat a\in\{\texttt{report},\texttt{abstain}\}.
\label{eq:intent-grounding}
\end{equation}
A valid report selects $\widehat S\subseteq\mathcal C$.
An explicit abstention selects $\widehat S=\varnothing$ and
does not initiate propagation. A structurally invalid
model response is recorded separately as a
\emph{rejected output}; it is not counted as a model-requested
abstention.
 
Let $G=(V,E,\tau_V,\tau_E)$ be the electrical traceability
graph, with node- and edge-type functions $\tau_V$ and
$\tau_E$. A deterministic resolver $\rho$ maps the selected
candidates to graph roots, and a depth-bounded typed
traversal $\Psi$ computes affected elements and paths:
\begin{equation}
\widehat R=\bigcup_{c\in\widehat S}\rho(c),
\qquad
\Psi(G,\widehat R,d)\rightarrow(\widehat A,\widehat P).
\label{eq:root-and-propagation}
\end{equation}
Here $d$ is the configured depth limit. Grounding is
evaluated against the expected action $a^*$ and intended
set $S^*$; root, affected-element, and typed-path outputs
are evaluated against reference sets $R^*$, $A^*$, and
$P^*$. This stage separation identifies whether an impact
error originates in candidate selection or downstream
propagation.
 
\subsection{Deterministic Candidate Extraction}
\label{subsec:candidate-extraction}
 
Revision differencing identifies additions, removals, and
attribute changes, then normalizes each atomic difference
into a candidate record. A candidate $c_i$ contains a
stable identifier $\mathrm{id}_i$, change type $t_i$, associated
entity identifiers $N_i$, observed operations $O_i$, and
optional semantic context $H_i$:
\begin{equation}
c_i=(\mathrm{id}_i,t_i,N_i,O_i,H_i).
\label{eq:candidate-record}
\end{equation}
An operation records its type, affected data path, and
before and after values. Depending on the scenario,
$H_i$ may contain functional names, registered aliases,
electrical endpoints, pins, harness and net identifiers,
requirement identifiers, and local typed relations.
Examples of observed changes include component
replacement, wire-gauge revision, terminal reassignment,
and interface modification.
 
Extraction does not use an LLM. The resulting inventory
includes request-relevant changes and genuine concurrent
distractors. Expected actions, intended candidates, and
reference graph outputs are kept apart from the information
available to grounding methods during prediction; the
evaluation protocol is specified in Section~\ref{sec:benchmark}.
 
\subsection{LLM Intent Grounding}
\label{subsec:llm-grounding}
 
Each LLM receives request $r$ and the model-visible
candidate inventory $\mathcal C$. It is instructed to
return a structured response containing
\texttt{action}, \texttt{selected\_candidate\_ids}, and
\texttt{rationale}. The requested action is
\texttt{report} or \texttt{abstain}. For a report, selected
identifiers must belong to $\mathcal C$; for an abstention,
the selected list must be empty. Response validation
prevents identifiers outside the inventory from becoming
graph roots. Rejected responses remain distinct from
explicit abstentions in the evaluation.
 
We evaluate Qwen2.5-7B, Llama3.1-8B, and Mistral-7B.
The model's task ends with candidate selection and action
prediction. It does not supply the final graph roots,
affected elements, or typed impact paths.
 
\subsection{Deterministic Graph Propagation}
\label{subsec:graph-propagation}
 
For a valid \texttt{report}, the resolver maps each
selected candidate to one or more graph roots. Roots may
represent components, wires, connectors, pins, harnesses,
nets, interfaces, or other traceable design entities.
Traversal follows permitted typed relations up to depth
$d$ and records both reached elements and the paths by
which they were reached. For fixed inputs, root resolution
and traversal are deterministic.
 
The oracle-root condition performs the same downstream
procedure using the reference intended candidates. It
provides a reference under perfect candidate selection;
it is not a deployable grounding method. Comparing this
condition with predicted-root propagation separates errors
introduced by grounding from the behavior of deterministic
graph replay.
 
\subsection{Compared Methods}
\label{subsec:compared-methods}
 
Table~\ref{tab:compared-methods} summarizes the comparison
conditions. Let $S_m$ be the valid candidate set selected
by model $m$, and let
$\mathcal M=\{\mathrm{Qwen},\mathrm{Llama},\mathrm{Mistral}\}$. For voting, an abstention or rejected response contributes no selected candidates. Define $D(r,c)=1$ when a distinguishing candidate-specific anchor in the model-visible record for $c$ is uniquely
supported by request $r$, and $D(r,c)=0$ otherwise.
The gate normalizes string-valued evidence from each
candidate record, excluding the candidate identifier. A
candidate is supported when the request contains a complete
normalized evidence string that occurs in only that
candidate's record. Such anchors may include entity
identifiers, aliases, functional names, or string-valued
before-and-after attributes. $D$ uses only $r$ and
model-visible candidate records; it never accesses $S^*$ or
other oracle annotations. The vote count
\begin{equation}
v(c)=\sum_{m\in\mathcal M}\mathbf 1[c\in S_m]
\label{eq:model-votes}
\end{equation}
includes all three models, including the base model when
its own consensus-adaptive output is evaluated.
 
\begin{table*}[t]
\centering
\caption{Intent-grounding and reference conditions in LOGIC. $\dagger$ denotes an exploratory post-hoc analysis.}
\label{tab:compared-methods}
\footnotesize
\begin{tabular}{p{0.24\textwidth}p{0.68\textwidth}}
\hline
\textbf{Condition} & \textbf{Selection rule or role} \\
\hline
All-difference & Select every observed candidate. \\
Token lexical & Select maximum-overlap candidates when the minimum token threshold is met. \\
Ungated LLM & Use an individual model's original validated action and selection. \\
Gate-only & Select candidates with request-matched string anchors that are unique within the inventory, without an LLM. \\
Uniform evidence gate & Retain only model-selected candidates with deterministic support. \\
Oracle-root & Propagate the intended candidates as a reference under perfect selection. \\
Majority ensemble$^\dagger$ & Select candidates receiving at least two of three model votes. \\
Consensus-adaptive$^\dagger$ & Retain a base model's selection if supported by evidence or model majority. \\
Evidence-empty abstention$^\dagger$ & Abstain when a model report has no supported selected candidate. \\
\hline
\end{tabular}
\end{table*}
 
The all-difference control ignores the request and selects
$S_{\mathrm{all}}=\mathcal C$. The token-lexical baseline
extracts normalized nontrivial token sets $T(r)$ and $T(c)$
and assigns $L(r,c)=|T(r)\cap T(c)|$. Let
$L_{\max}=\max_{c\in\mathcal C}L(r,c)$. It selects all
candidates tied at $L_{\max}$ when $L_{\max}\ge 2$ and
otherwise abstains. This comparison tests surface-form
alignment. The ungated condition uses the individual
model's original validated selection and action.
 
Gate-only selects
$S_{\mathrm{gate}}=\{c\in\mathcal C:D(r,c)=1\}$.
The uniform evidence gate instead filters an individual
model's selection:
\begin{equation}
S_{\mathrm{uniform},m}
=\{c\in S_m:D(r,c)=1\}.
\label{eq:uniform-gate}
\end{equation}
It never adds a candidate. A model abstention or rejected
output is preserved; a report becomes an abstention if no
selected candidates survive filtering.
 
The majority ensemble and consensus-adaptive policy use
the vote count in~\eqref{eq:model-votes}:
\begin{align}
S_{\mathrm{majority}}
&=\{c\in\mathcal C:v(c)\geq 2\},
\label{eq:majority}\\
S_{\mathrm{adaptive},m}
&=\{c\in S_m:D(r,c)=1\ \lor\ v(c)\geq 2\}.
\label{eq:adaptive-gate}
\end{align}
Consensus-adaptive gating cannot add a candidate absent
from its base model's selection. It requires predictions
from all three models and is therefore an ensemble policy.
Consensus-adaptive filtering preserves an original abstention or rejected output, a valid report becomes an abstention when no candidates remain. For the majority ensemble, an empty selection yields \texttt{abstain}.
 
Evidence-empty abstention acts at the request level. For
an original model report, it changes the action to
\texttt{abstain} if
$\sum_{c\in S_m}D(r,c)=0$; otherwise, it retains the
original action and selection. An original abstention or
rejected output is preserved.
 
Finally, let $\pi(r,\mathcal C)=\mathbf{1}[\exists c\in\mathcal C:D(r,c)=1]$
indicate whether any inventory candidate has unique
deterministic support. The binary diagnostic predicts
\emph{literal} when $\pi(r,\mathcal C)=1$ and \emph{nonliteral}
otherwise. For the exploratory three-way diagnostic, let
$A_{\mathrm{vote}}=\mathbf{1}[\exists c\in\mathcal C:v(c)\geq 2]$.
It predicts \emph{literal} if $\pi(r,\mathcal C)=1$,
\emph{semantic} if $\pi(r,\mathcal C)=0$ and
$A_{\mathrm{vote}}=1$, and \emph{safety} otherwise. These are
diagnostic labels, not an automatic reporting or abstention
policy.
\section{LOGIC Benchmark and Experimental Design}
\label{sec:benchmark}

\begin{figure*}[t]
\centering
\includegraphics[width=\textwidth]{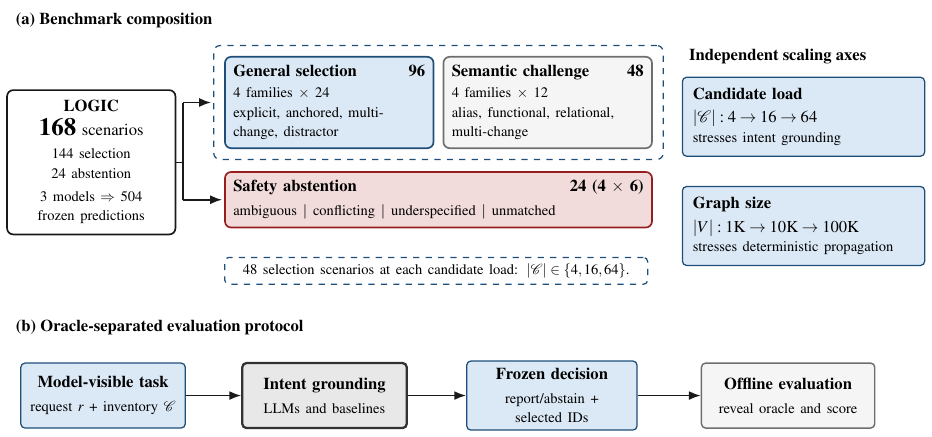}
\caption{LOGIC benchmark and evaluation protocol. The
168 scenarios comprise 144 answerable selection requests
and 24 abstention requests. Candidate count varies the
model-visible grounding task; graph-size replay evaluates
deterministic propagation from frozen selections. Three
local LLMs produce 504 predictions before oracle-based
offline scoring.}
\label{benchmark_protocol}
\end{figure*}

LOGIC contains 168 controlled scenarios: 144 requests for
which at least one observed change should be selected and
24 requests for which the expected action is
\texttt{abstain}. Figure~\ref{benchmark_protocol}
summarizes the scenario organization and evaluation
protocol. The scenarios are designed to vary request
evidence, distractor similarity, candidate load, and graph
size under known reference answers. Their proportions are
experimental design choices, not estimates of how often
these conditions occur in operational aerospace work.

\subsection{Scenario Construction and Request Families}
\label{subsec:benchmark-composition}

Each scenario pairs a natural-language engineering change
request with an inventory of genuine changes between
synthetic before and after ECAD revisions. Controlled
revision operations create the intended changes and
concurrent distractors; deterministic differencing
produces the candidate records supplied to grounding
methods. Reference actions and intended candidate sets
are recorded separately. Fictional identifiers and design values provide a controlled, reproducible representation of aerospace electrical revisions.

The 144 answerable requests contain 96 general
controlled-selection cases and 48 targeted semantic
challenges. General controlled selection has four families
of 24 cases each: \emph{explicit single change} identifies
one change through a direct anchor; \emph{multi-change}
requests require a set of intended changes;
\emph{paraphrased functional} requests describe a function
while retaining recoverable structured evidence; and
\emph{same-entity distractor} cases require distinguishing
one operation from other changes involving the same or a
closely related entity.

The targeted semantic challenges have four families of
12 cases each. \emph{Exact alias} provides a registered
literal alias; \emph{functional paraphrase} describes an
engineering function in different words;
\emph{relational paraphrase} requires interpreting
endpoints, connectivity, or local typed relations; and
\emph{semantic multi-change} requests describe more than
one intended change through functional or relational
language. Exact-alias cases provide a direct-evidence
reference within this group. The remaining families
concentrate conditions in which literal evidence can be
incomplete. These families form a deliberately balanced diagnostic sample for evaluating semantic grounding rather than a sample of operational request frequencies.

The 24 abstention requests comprise six cases each of
\emph{ambiguous}, \emph{conflicting},
\emph{underspecified}, and \emph{unmatched} intent.
An ambiguous request is consistent with multiple
candidates without identifying one uniquely; a
conflicting request contains incompatible selection
evidence; an underspecified request omits needed
distinguishing attributes; and an unmatched request has
no implementing candidate in the inventory. Their
reference action is \texttt{abstain}.

Scenario generation and checking are deterministic.
Candidate operations are checked against the underlying
before and after designs, and the reference candidate
identifiers must belong to the generated inventory.
Model-visible records are checked for reference-label
fields. These checks establish internal consistency; independent domain-expert assessment of engineering realism remains future work.

\subsection{Candidate Load and Graph Replay}
\label{subsec:controlled-scaling}

Candidate inventories contain 4, 16, or 64 detected
changes. There are 48 answerable scenarios at each
candidate count: 32 general controlled-selection and
16 targeted semantic cases. Increasing candidate count
adds genuine competing revision changes to the
model-visible inventory. In the general cases, target
positions and distractor similarity are also controlled
to avoid treating one inventory order or one easy
distractor pattern as representative.

Graph size is varied independently at approximately 1K,
10K, and 100K nodes. The evaluated synthetic designs
contain components and pins, wires, requirements, and
verification activities. Typed relations represent
component--pin containment, wire endpoints and connected
components, requirement allocation or traceability, and
verification links. Candidate metadata can additionally
describe functional names, aliases, harnesses, nets, and
local relationships. The full propagation graph is not
included in the LLM prompt.

For graph replay, already frozen candidate selections are
resolved to roots in each graph and traversed using the
same typed, depth-bounded procedure. The request and
selection are not regenerated as graph size changes.
Candidate-count results therefore measure how grounding
accuracy responds to a larger model-visible inventory;
graph-size results measure the stability of downstream
root, affected-element, and typed-path accuracy. They do
not establish traversal-time scalability.

The request specifies a maximum traversal depth, and the
reported experiments use a depth limit of $d=3$. Prompt
audits check the serialized model-visible request and
inventory against the configured context limit; the audit
outcome and prompt-generation procedure are retained with
the benchmark artifacts.

\subsection{Model Inference and Oracle Separation}
\label{subsec:oracle-separation}

The model-input package contains the request, candidate
inventory, scenario identifier, and configuration
metadata required to build the prompt. A separate oracle
package contains the expected action, intended candidate
identifiers, reference graph roots, affected elements,
and typed paths. Oracle fields are unavailable during
model prediction and deterministic baseline selection.

Qwen2.5-7B, Llama3.1-8B, and Mistral-7B each produce
one prediction per scenario, yielding 504 frozen model
predictions. The models run locally under a fixed
structured-response contract. The frozen prediction
artifacts record raw responses, parsed actions and
candidate identifiers, validation diagnostics, model
metadata, prompt and input hashes, and timing data.
All three models were run with temperature 0, a
32,768-token configured context limit, and a 300-token
generation limit. The recorded run seed was 2609 for the
120 general and safety cases and 2709 for the 48 targeted
semantic cases, with the same seed used across all three
models within each split.

Model predictions and deterministic baseline selections are fixed before offline scoring against separately stored reference annotations. Scoring does not edit a
selection or regenerate model responses using reference information. Token-lexical and gate-only baselines are evaluated across all 168 scenarios using the same model-visible inputs as the LLMs; ensemble and abstention analyses reuse the frozen model predictions. The majority-vote, consensus-adaptive, evidence-empty
abstention, and regime-detection analyses were developed
after examining primary benchmark behavior and are
reported as exploratory.

\subsection{Evaluation Metrics and Statistical Analysis}
\label{subsec:evaluation-metrics}

Candidate-selection metrics are computed on the 144
answerable scenarios. For predicted set $\widehat{S}$ and
reference set $S^{*}$, true positives, false positives, and
false negatives are aggregated across scenarios:
\begin{equation}
\begin{aligned}
TP&=|\widehat S\cap S^*|,&
FP&=|\widehat S\setminus S^*|,\\
FN&=|S^*\setminus\widehat S|.&
\end{aligned}
\label{eq:candidate-counts}
\end{equation}
The reported candidate precision and recall are
$P_{\mathrm{cand}}=TP/(TP+FP)$ and
$R_{\mathrm{cand}}=TP/(TP+FN)$, with candidate F1 their
harmonic mean. Exact-set accuracy is the fraction of
answerable scenarios for which $\widehat{S}=S^{*}$.
Action accuracy compares the predicted
\texttt{report} or \texttt{abstain} action with the
reference action.

On the 24 safety scenarios, \emph{strict abstention
accuracy} counts only valid explicit \texttt{abstain}
outputs as correct. The \emph{unsafe-report rate} counts
\texttt{report} outputs where abstention is expected.
Malformed or otherwise rejected outputs are recorded
separately and receive no credit as correct abstentions.

On answerable scenarios, \emph{coverage} is the fraction
receiving a valid \texttt{report}, while the
\emph{false-abstention rate} is the fraction receiving
\texttt{abstain}. Rejected outputs are reported separately. For evidence-state analyses, a selected candidate is \emph{supported} when $D(r,c)=1$ under the
deterministic rule in Section~\ref{subsec:compared-methods};
otherwise it is \emph{unsupported}. Correctness is
determined against $S^{*}$ only during offline scoring.

Root, affected-element, and typed-path sets are scored
separately. For $X\in\{R,A,P\}$, set-level F1 is
\begin{equation}
F1_X=
\frac{2|\widehat X\cap X^*|}
{|\widehat X|+|X^*|}.
\label{eq:graph-f1}
\end{equation}
The oracle-root condition supplies the reference
selection to the same deterministic propagation pipeline.
Scores of 1.0 under this condition are expected by
construction for the generated graph references.

For paired method comparisons, scenarios are resampled
as paired units to estimate 95\% bootstrap confidence
intervals for differences in candidate F1, exact-set
accuracy, abstention accuracy, and coverage. Binary
scenario-level correctness comparisons use exact
McNemar tests. Holm adjustment is applied within each
reported family of multiple comparisons. Semantic-family
results are interpreted cautiously because each family
contains only 12 scenarios. Regime-detector accuracy and
macro-F1 are computed over the benchmark's labeled
request regimes.

\section{Results}
\label{sec:results}

We report results from the controlled 168-scenario LOGIC benchmark. Candidate-selection performance is evaluated on 144 answerable cases, comprising 96 general controlled-selection cases and 48 targeted semantic challenges. Abstention is evaluated separately on 24 safety cases. Candidate precision, recall, and F1 use pooled candidate counts within the reported group; exact-set accuracy and action outcomes are measured per scenario.

\subsection{RQ1: How accurately do the methods ground engineering requests?}
\label{subsec:rq1-grounding}

Table~\ref{tab:grounding-load} combines overall candidate selection with performance at each inventory size. On the 96 general controlled-selection cases, all-difference obtains perfect recall but only 0.0446 precision, producing F1 of 0.0855 and no exact candidate sets. Gate-only achieves precision, recall, F1, and exact-set accuracy of 1.0000. Token lexical remains strong, with precision of 0.9375, recall of 1.0000, F1 of 0.9677, and exact-set accuracy of 0.9583. Its eight false-positive selections occur in four multi-change cases at the largest candidate load.

Ungated performance varies across the three LLMs. Qwen2.5-7B achieves F1 of 0.7791, Llama3.1-8B achieves 0.7698, and Mistral-7B achieves 0.6639. Uniform evidence gating raises these values to 0.8940, 0.8940, and 0.8000, respectively. Three-model majority voting obtains F1 of 0.8190. Neither the evaluated LLM conditions nor majority voting exceeds either deterministic baseline in this group.

\begin{figure*}[t]
\centering
\includegraphics[width=\textwidth]{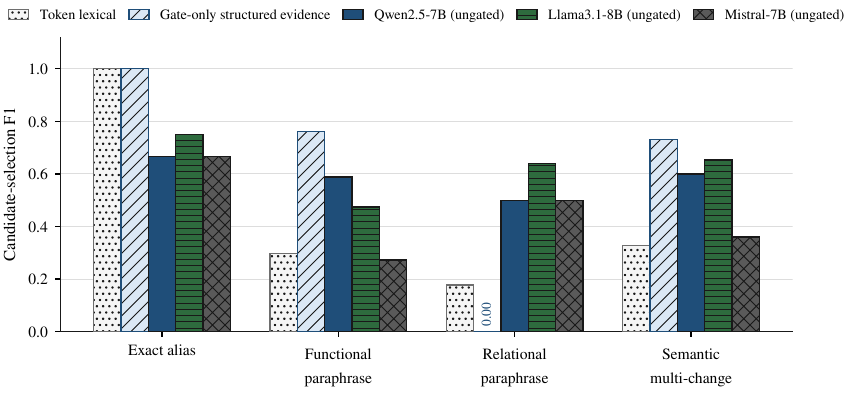}
\caption{Candidate-selection F1 across the four semantic-challenge families, each containing 12 cases. Ungated LLMs exceed token lexical and gate-only on relational paraphrases; gate-only performs better on functional paraphrases and semantic multi-change requests. Family-level exact-set comparisons do not remain significant after Holm correction.}
\label{fig:semantic_families}
\end{figure*}

The semantic challenges expose a different pattern. Token lexical and gate-only both achieve exact selection on exact-alias requests, but their behavior differs across the other families (Fig.~\ref{fig:semantic_families}). On relational paraphrases, token-lexical F1 is 0.1772 and gate-only F1 is 0.0000, whereas ungated Qwen and Mistral each achieve 0.5000 and Llama achieves 0.6400. The paired-bootstrap F1 differences from token lexical are $+0.3228$ for Qwen (95\% CI $[+0.0340,+0.5609]$), $+0.4628$ for Llama ($[+0.1664,+0.7460]$), and $+0.3228$ for Mistral ($[+0.0698,+0.5540]$).

On the 12 relational-paraphrase cases, the reported bootstrap intervals support an F1 advantage over token lexical, although exact-set differences are not statistically significant after Holm correction. Gate-only exceeds every ungated LLM on functional paraphrases and semantic multi-change requests, achieving F1 of 0.7619 and 0.7317, respectively. These results indicate complementary strengths that the evaluated LLMs handle some relational descriptions better, while deterministic evidence remains effective when distinctive literal anchors are available.

\begin{table*}[t]
\caption{Candidate-selection accuracy and sensitivity to candidate load. Overall F1 and exact-set accuracy are computed over the answerable cases in each group; the final three columns report F1 by inventory size. Model-independent methods are shown once and marked ``---'' in the Model(s) column. Dashes in numerical columns indicate omitted breakdowns. $\dagger$ denotes an exploratory ensemble analysis.}
\label{tab:grounding-load}
\centering
\small
\setlength{\tabcolsep}{5pt}
\begin{tabular}{llccccc}
\hline
\textbf{Method} & \textbf{Model(s)} & \textbf{Overall F1} & \textbf{Exact}
& $\boldsymbol{|\mathcal C|=4}$ & $\boldsymbol{|\mathcal C|=16}$ & $\boldsymbol{|\mathcal C|=64}$ \\
\hline
\multicolumn{7}{l}{\textbf{(a) General controlled selection: 96 cases, 32 per inventory size}} \\
\hline
All-difference & --- & 0.0855 & 0.0000 & -- & -- & -- \\
Token lexical & --- & 0.9677 & 0.9583 & 1.0000 & 1.0000 & 0.9091 \\
Gate-only & --- & 1.0000 & 1.0000 & 1.0000 & 1.0000 & 1.0000 \\
Oracle-root & --- & 1.0000 & 1.0000 & 1.0000 & 1.0000 & 1.0000 \\
Ungated LLM & Qwen & 0.7791 & 0.7083 & 1.0000 & 0.8222 & 0.5063 \\
& Llama & 0.7698 & 0.7188 & 0.9750 & 0.7273 & 0.6190 \\
& Mistral & 0.6639 & 0.5729 & 0.8861 & 0.6517 & 0.4384 \\
Uniform gate & Qwen & 0.8940 & 0.7812 & 1.0000 & 0.9610 & 0.6667 \\
& Llama & 0.8940 & 0.8021 & 0.9873 & 0.8889 & 0.7879 \\
& Mistral & 0.8000 & 0.6458 & 0.9333 & 0.8406 & 0.5714 \\
Majority vote$^\dagger$ & Three models & 0.8190 & 0.6875 & -- & -- & -- \\
\hline
\multicolumn{7}{l}{\textbf{(b) Targeted semantic challenges: 48 cases, 16 per inventory size}} \\
\hline
All-difference & --- & 0.0855 & 0.0000 & -- & -- & -- \\
Token lexical & --- & 0.3346 & 0.3750 & 0.7317 & 0.4110 & 0.1818 \\
Gate-only & --- & 0.6931 & 0.5208 & 0.7317 & 0.6667 & 0.6667 \\
Oracle-root & --- & 1.0000 & 1.0000 & 1.0000 & 1.0000 & 1.0000 \\
Ungated LLM & Qwen & 0.5905 & 0.4375 & 0.9000 & 0.5128 & 0.2308 \\
& Llama & 0.6176 & 0.6250 & 1.0000 & 0.5098 & 0.4000 \\
& Mistral & 0.4085 & 0.4375 & 0.9268 & 0.2414 & 0.1395 \\
Uniform gate & Qwen & 0.4938 & 0.3125 & 0.7647 & 0.5185 & 0.0000 \\
& Llama & 0.6813 & 0.4792 & 0.8571 & 0.5714 & 0.5714 \\
& Mistral & 0.4938 & 0.3333 & 0.8333 & 0.2609 & 0.1818 \\
Majority vote$^\dagger$ & Three models & 0.6168 & 0.5417 & -- & -- & -- \\
\hline
\end{tabular}
\end{table*}

\subsection{RQ2: How does candidate load affect LLM grounding?}
\label{subsec:rq2-candidate-scaling}

The rightmost columns of Table~\ref{tab:grounding-load} show that all three ungated LLMs perform best with four candidates and deteriorate as the inventory grows. On the general cases, the largest absolute decline is Qwen's change from F1 of 1.0000 to 0.5063 between four and 64 candidates. Llama retains the highest ungated F1 at 64 candidates, at 0.6190. Uniform gating improves performance at the larger loads but does not eliminate the decline. Gate-only remains at F1 of 1.0000 at every load. Token lexical also achieves 1.0000 at four and 16 candidates, but falls to 0.9091 at 64 candidates through additional false-positive selections.

The semantic challenges amplify the difficulty. Between four and 64 candidates, ungated F1 falls by 0.6692 for Qwen, 0.6000 for Llama, and 0.7873 for Mistral. At four candidates, each LLM exceeds the deterministic baselines; at 64, gate-only exceeds each ungated model. These results identify candidate-inventory growth as a central challenge for LLM intent grounding, the models perform strongly on small inventories, while gate-only is more robust at the largest tested load.

\subsection{RQ3: When does deterministic evidence help or suppress LLM selections?}
\label{subsec:rq3-evidence}

\begin{figure*}[t]
\centering
\includegraphics[width=0.94\textwidth]{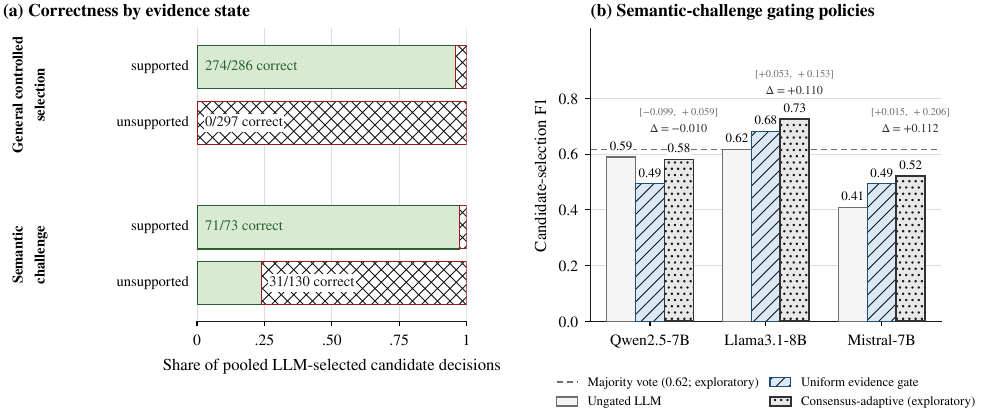}
\caption{Deterministic evidence and exploratory gating. (a) Correctness of supported and unsupported ungated selections, pooled across three models. The general-and-safety group includes 120 cases; the semantic-challenge group includes 48. (b) Semantic F1 for ungated, uniformly gated, and consensus-adaptive selections. The dashed line marks three-model majority-vote F1. Printed values give the adaptive-minus-ungated F1 difference with its paired-bootstrap 95\% confidence interval.}
\label{fig:evidence_gating}
\end{figure*}

Figure~\ref{fig:evidence_gating}(a) groups individual candidate selections from the three ungated LLMs by deterministic support and reference correctness. Counts are pooled across models and scenarios, so the same candidate selected by different models is counted separately. Across the 120 general and safety cases, 274 of 286 supported selections are correct, compared with none of the 297 unsupported selections. These counts include the 24 cases requiring abstention, where any selected candidate is incorrect by the reference action. They therefore describe the combined selection-and-safety evaluation rather than a positive-selection-only estimate. The strong separation also reflects the explicit-anchor construction of the general cases.

In the 48 semantic challenges, 71 of 73 supported selections are correct, but 31 of 130 unsupported selections are also correct: 11 of 24 for Qwen, 11 of 45 for Llama, and 9 of 61 for Mistral. Uniform gating removes many false positives while discarding valid selections for which its rules provide no support. Consequently, absence of rule-recognized evidence is informative but does not establish that a semantic interpretation is wrong.

We evaluated cross-model consensus-adaptive gating as an exploratory response to this tradeoff. Starting from one model's selection, it retains a candidate when deterministic evidence supports it or at least two of the three models selected it. It never adds a candidate omitted by the base model. On the semantic challenges, F1 changes from 0.5905 to 0.5806 for Qwen, from 0.6176 to 0.7273 for Llama, and from 0.4085 to 0.5200 for Mistral. The paired-bootstrap differences from ungated output are $-0.0098$ (95\% CI $[-0.0988,+0.0589]$), $+0.1096$ ($[+0.0525,+0.1531]$), and $+0.1115$ ($[+0.0150,+0.2058]$), respectively. Qwen's recall decreases from 0.5167 to 0.4500, illustrating that consensus filtering can remove correct selections as well as errors.

Majority voting provides the corresponding ensemble baseline. It achieves semantic F1 of 0.6168 and exact-set accuracy of 0.5417. All 18 unanimously selected semantic candidates are correct, whereas candidates selected by only one model comprise 18 correct and 73 incorrect selections. Consensus-adaptive Llama exceeds majority-vote F1 by 0.1105 (95\% CI $[+0.0399,+0.1960]$), while consensus-adaptive Mistral falls below it by 0.0968 ($[-0.1870,-0.0274]$). Adaptive Qwen also has lower point-estimate F1 than majority voting. The paired F1 intervals identify model-dependent gains and losses, while exact-set comparisons remain nonsignificant after Holm correction. Consensus-adaptive gating requires all three models, and its benefit depends on the base model's error pattern.

\subsection{RQ4: Can evidence-informed abstention reduce unsafe reporting?}
\label{subsec:rq4-abstention}

Table~\ref{tab:safety-results}(a) reports outcomes on the 24 safety cases together with coverage on the 144 answerable cases. Gate-only correctly abstains in 18 safety cases and reports unsafely in six, whereas token lexical correctly abstains in nine and reports unsafely in 15. Ungated Qwen correctly abstains in ten cases; Llama and Mistral correctly abstain in none. Each ungated model produces four rejected outputs, which are recorded separately from deliberate abstentions.

The exploratory evidence-empty policy changes a model report to abstention when none of its selected candidates has unique deterministic support. It raises strict abstention accuracy to 16 of 24 (0.6667) and reduces unsafe reports to four of 24 (0.1667) for every model. The uniform evidence gate produces the same safety-outcome counts, although it additionally filters candidate selections on answerable cases. Gate-only has higher strict abstention accuracy than the evidence-informed LLM policies, but those policies produce fewer unsafe reports and retain four rejected outcomes per model.

Relative to ungated output, evidence-empty abstention improves strict accuracy by 0.2500 for Qwen (paired-bootstrap 95\% CI $[+0.0833,+0.4167]$) and 0.6667 for both Llama and Mistral ($[+0.4583,+0.8333]$ for each). Holm-adjusted exact McNemar tests yield $p=0.03125$ for Qwen and $p<0.001$ for Llama and Mistral. The improvement costs 16.0, 25.0, and 27.1 percentage points of positive-case coverage, respectively. Exact positive-case accuracy also decreases by 4.17, 5.56, and 4.17 percentage points.

The family breakdown in Table~\ref{tab:safety-results}(b) shows where this policy succeeds and fails. Evidence-empty abstention suppresses unsafe reporting on ambiguous, underspecified, and unmatched requests. Four of six conflicting requests nevertheless remain unsafe for every model. Thus, evidence absence offers a useful abstention signal, but evidence presence does not resolve contradictory instructions.

\begin{table*}[t]
\caption{Safety and coverage outcomes. Panel (a) reports rates on 24 safety cases and report coverage on 144 answerable cases. Panel (b) reports \emph{correct-abstention / unsafe-report counts} within each six-case family; rejected outputs account for any remainder. $\dagger$ denotes an exploratory policy.}
\label{tab:safety-results}
\centering
\small
\setlength{\tabcolsep}{5pt}
\begin{tabular}{llcccc}
\hline
\multicolumn{6}{l}{\textbf{(a) Overall safety outcomes and positive-case coverage}} \\
\hline
\textbf{System} & \textbf{Policy} & \textbf{Strict abst.} & \textbf{Unsafe} & \textbf{Rejected} & \textbf{Coverage} \\
\hline
--- & Token lexical & 0.3750 & 0.6250 & 0.0000 & 1.0000 \\
--- & Gate-only & 0.7500 & 0.2500 & 0.0000 & 0.8750 \\
Qwen & Ungated & 0.4167 & 0.4167 & 0.1667 & 0.8333 \\
& Evidence-empty$^\dagger$ & 0.6667 & 0.1667 & 0.1667 & 0.6736 \\
Llama & Ungated & 0.0000 & 0.8333 & 0.1667 & 0.9931 \\
& Evidence-empty$^\dagger$ & 0.6667 & 0.1667 & 0.1667 & 0.7431 \\
Mistral & Ungated & 0.0000 & 0.8333 & 0.1667 & 0.8681 \\
& Evidence-empty$^\dagger$ & 0.6667 & 0.1667 & 0.1667 & 0.5972 \\
Three models & Majority vote$^\dagger$ & 0.4583 & 0.5417 & 0.0000 & 0.8333 \\
\hline
\multicolumn{6}{l}{\textbf{(b) Safety-family outcome counts: six cases per family}} \\
\hline
\textbf{System} & \textbf{Policy} & \textbf{Ambiguous} & \textbf{Conflicting} & \textbf{Underspecified} & \textbf{Unmatched} \\
\hline
--- & Token lexical & 3/3 & 0/6 & 6/0 & 0/6 \\
--- & Gate-only & 6/0 & 0/6 & 6/0 & 6/0 \\
Qwen & Ungated & 0/4 & 0/6 & 4/0 & 6/0 \\
& Evidence-empty$^\dagger$ & 4/0 & 2/4 & 4/0 & 6/0 \\
Llama & Ungated & 0/4 & 0/6 & 0/4 & 0/6 \\
& Evidence-empty$^\dagger$ & 4/0 & 2/4 & 4/0 & 6/0 \\
Mistral & Ungated & 0/5 & 0/5 & 0/4 & 0/6 \\
& Evidence-empty$^\dagger$ & 5/0 & 1/4 & 4/0 & 6/0 \\
Three models & Majority vote$^\dagger$ & 2/4 & 2/4 & 2/4 & 5/1 \\
\hline
\end{tabular}
\end{table*}

\subsection{Exploratory Regime Detection}
\label{subsec:regime-results}

The inventory-level evidence-presence diagnostic reaches accuracy of 0.8571 and macro-F1 of 0.8250 across 168 cases. Literal recall and nonliteral precision are both 1.0000, but nonliteral recall is 0.6000 because 24 of 60 nonliteral requests contain some deterministic support. The three-way evidence-and-consensus diagnostic is weaker, with accuracy of 0.7679 and macro-F1 of 0.5942. Evidence absence is therefore a precise but incomplete signal for nonliteral requests within LOGIC; evidence presence alone cannot determine whether a request is literal, contradictory, or safe to report.

\subsection{RQ5: How stable is propagation accuracy as graph size increases?}
\label{subsec:rq5-propagation}

Table~\ref{tab:graph-replay} reports deterministic replay of frozen ungated selections over approximately 1K-, 10K-, and 100K-node graphs for the semantic-challenge cases. Root F1 is invariant by construction because the same selected candidates resolve to the same logical roots before each replay. Its values are 0.4833 for Qwen, 0.6796 for Llama, and 0.4851 for Mistral.

Affected-element and typed-path accuracy vary modestly over the tested graph sizes. Llama's affected-element F1 changes from 0.724 at 1K nodes to 0.699 at 100K, while its typed-path F1 changes from 0.652 to 0.639. This contrasts with the substantially larger candidate-load declines. Under the tested replay protocol and depth limit of three, the larger measured accuracy limitation lies in selecting the intended roots from competing changes. These measurements establish graph-size accuracy stability; they do not establish traversal-time scalability.

\begin{table*}[t]
\caption{Deterministic graph-replay accuracy from frozen ungated selections on the 48 semantic challenges. Root F1 is constant across sizes and is reported in the text.}
\label{tab:graph-replay}
\centering
\begin{tabular}{lcccccc}
\hline
& \multicolumn{3}{c}{\textbf{Affected-element F1}}
& \multicolumn{3}{c}{\textbf{Typed-path F1}} \\
\cline{2-7}
\textbf{Model} & \textbf{1K} & \textbf{10K} & \textbf{100K}
& \textbf{1K} & \textbf{10K} & \textbf{100K} \\
\hline
Qwen2.5-7B & 0.505 & 0.507 & 0.504 & 0.474 & 0.473 & 0.478 \\
Llama3.1-8B & 0.724 & 0.726 & 0.699 & 0.652 & 0.639 & 0.639 \\
Mistral-7B & 0.547 & 0.539 & 0.515 & 0.466 & 0.463 & 0.466 \\
Oracle-root & 1.000 & 1.000 & 1.000 & 1.000 & 1.000 & 1.000 \\
\hline
\end{tabular}
\end{table*}

\section{Discussion}
\label{sec:discussion}

The results support four observations, discussed in turn below. The LLM advantage is specific to relational requests, while deterministic methods remain preferable when identifying evidence is present. Support-based gating trades recall for precision because absent evidence is not contradiction. Abstention reduces unsafe reports at a measurable coverage cost and must remain distinct from rejection and from a ``no impact'' result. Finally, separating grounding from deterministic propagation localizes the observed scaling limitation to candidate selection, although it does not establish operational readiness.

\subsection{Complementary Grounding Capabilities}
\label{subsec:discussion-complementary}

LOGIC identifies conditions under which local LLMs contribute to intent grounding. Literal identifiers and directly matching attributes favor deterministic methods: gate-only achieves exact selection on all general controlled-selection cases, while token lexical achieves F1 of 0.9677 and exact-set accuracy of 0.9583. Gate-only also exceeds the ungated models on functional paraphrases and semantic multi-change requests. The observed LLM advantage over both deterministic methods is concentrated in the 12 relational-paraphrase cases, where candidate F1 remains between 0.5000 and 0.6400. 

The practical implication is to preserve reliable deterministic grounding while investigating LLM assistance for requests whose relationships are not captured by the available rules. However, evidence presence alone cannot reliably determine when literal matching is sufficient, because relational and conflicting requests may also contain recognizable identifiers or attributes. A hybrid architecture therefore needs a mechanism for distinguishing applicable evidence from incidental or contradictory matches.

\subsection{Evidence Support and the Precision--Recall Tradeoff}
\label{subsec:discussion-evidence}

Uniform gating exposes a limitation of support-based validation. A candidate may lack a recognized identifier or anchor while still satisfying a relational request. Consequently, absence of deterministic support does not imply deterministic contradiction. The gate's precision gains must be assessed together with the correct selections it removes, particularly when the rules cover only part of the request semantics.

A useful future distinction is between \emph{supported}, \emph{contradicted}, and \emph{unresolved} selections. Support indicates positive grounding evidence; contradiction requires an explicit inconsistency with the request or another constraint; unresolved indicates that the implemented checks cannot decide. LOGIC evaluates support-based filtering, but does not validate a complete contradiction detector. This three-state formulation is therefore a proposed extension.

Cross-model agreement provides another signal, with mixed results. Consensus-adaptive filtering improves semantic F1 for Llama and Mistral relative to their ungated outputs, but only adaptive Llama exceeds the majority-vote baseline. Qwen loses recall without a net F1 gain. These exploratory results show that agreement-based filtering depends on the base model's error pattern. It also requires three model outputs, whose agreement should not be interpreted as independent evidence or calibrated confidence.

\subsection{Abstention, Rejection, and Engineering Review}
\label{subsec:discussion-safety}

Evidence-empty abstention reduces unsafe reporting by moving some requests out of automatic processing. Its coverage loss is therefore part of the policy's operating cost, rather than an incidental drawback. The remaining failures on conflicting requests show that a supported selection can still be inappropriate when the request contains incompatible constraints. Conflicting requests therefore require explicit consistency checks and engineering review before reporting.

An engineering interface should distinguish three outcomes:
\begin{itemize}
    \item \emph{Report}: a valid candidate selection initiates deterministic propagation, with the selected changes and resulting paths available for review.
    \item \emph{Abstain}: the grounding method deliberately declines to select changes and surfaces the unresolved request.
    \item \emph{Reject}: response validation finds the model output unusable, for example because it violates the response contract.
\end{itemize}

Neither abstention nor rejection should become an empty ``no impact'' report. Such a report could imply that downstream consequences were evaluated when the initiating changes were never established. Review records should instead preserve the request, candidate inventory, available model output, and reason that processing stopped. In this setting, LLM grounding supports engineering assessment; it does not authorize a design change or replace the responsible engineer's decision.

\subsection{Deployment Implications}
\label{subsec:discussion-deployment}

The evaluated 7--8B models support local execution, allowing design records and predictions to remain within organizational infrastructure. This makes the approach relevant to engineering environments with restrictions on transmitting proprietary information to external services. Local execution also permits model artifacts and inference settings to be recorded as part of a reproducible configuration, although the study does not validate an operational air-gapped deployment.

The separation between model-based selection and deterministic graph propagation provides a clear boundary for evaluation and review. Candidate-load experiments reveal sensitivity in the grounding stage, while graph replay shows comparatively stable downstream accuracy for fixed selections over the tested sizes. These are distinct experimental axes and do not establish relative computational scalability. Every reported downstream element can nevertheless be traced to a selected candidate and an explicit typed path.

Reducing the model-visible inventory through retrieval or hierarchical filtering is a potential response to candidate-load sensitivity. Any such prefilter must retain the intended changes: a candidate removed before grounding cannot be recovered by the LLM. Evaluating that recall constraint, together with representative engineering cases and measured execution costs, is a necessary next step.

\section{Threats to Validity}
\label{sec:threats}

\subsection{Benchmark Construction and Ecological Validity}
\label{subsec:threats-construction}

LOGIC evaluates LLM intent grounding using synthetic engineering requests, revision-change inventories, and traceability graphs developed within the project. Controlled construction provides exact reference answers and genuine concurrent distractors, but does not reproduce the full variability of industrial ECAD revisions. Organization-specific terminology, incomplete records, legacy identifiers, and information distributed across drawings or engineering systems may change grounding behavior. Independent subject-matter-expert validation was not part of the reported evaluation.

The general controlled-selection cases contain recoverable literal or structured anchors, producing a gate-only ceiling of 1.0000 and token-lexical F1 of 0.9677. These results reflect the construction of those cases and should not be generalized to operational requests. The semantic challenges provide additional contextual fields, including functional names, endpoints, and local relations. Consequently, differences between these groups reflect both request formulation and candidate representation; within-group comparisons use the same model-visible records.

Synthetic graph size also does not establish representative topology. The replay experiments assess accuracy under the tested relationships and depth limit, rather than industrial coverage, traversal-time scalability, or end-to-end processing cost; traversal latency, memory consumption, extraction cost, and report-generation time require separate evaluation.

\subsection{Diagnostic Construction and Evaluation Integrity}
\label{subsec:threats-diagnostic}

The 48 semantic-challenge cases were constructed after observing limitations of deterministic evidence gating. They are therefore a targeted diagnostic slice of LOGIC rather than an independently sampled confirmatory test set. Their balanced family composition supports controlled comparisons but does not estimate the prevalence of relational or other nonliteral requests in engineering practice.

Run artifacts record the prompt and evidence-rule specifications used during prediction. Model-visible inputs were separated from oracle references, and predictions were frozen and hash-verified before offline scoring. These controls reduce direct label exposure and preserve the executed experiment, but do not remove construction bias or regularities shared by the benchmark generator and grounding methods. They also do not make subsequent diagnostic experiments independent of earlier observations.

\subsection{Model Coverage and Statistical Uncertainty}
\label{subsec:threats-statistical}

The evaluation covers three local 7--8B models, one inference configuration per run, and one frozen prediction per model and scenario. It does not establish robustness across model sizes, alternative prompts, quantization choices, or repeated executions. The two scenario groups also used different recorded seeds, and their representation differences limit attribution of cross-group performance changes solely to request semantics.

Each semantic family contains 12 cases and each safety family contains six. These small samples limit paired-test power, particularly after multiple-comparison correction. Bootstrap intervals describe uncertainty under the chosen resampling procedure, but cases that share a request template across candidate loads may be correlated in ways that scenario-level resampling does not fully capture. Family findings should therefore be interpreted as diagnostic evidence with limited population generalizability.

Candidate evidence counts pool selections across models and requests. They are descriptive totals, not independent observations: the models evaluate the same cases, and multiple candidates may arise from one request. Similarly, unanimous model agreement is not calibrated confidence or evidence of independent reasoning.

\subsection{Exploratory Policy Evaluation}
\label{subsec:threats-exploratory}

Majority voting, consensus-adaptive gating, evidence-empty abstention, and regime detection were examined after observing primary benchmark behavior. Applying them to frozen outputs allows controlled reanalysis without changing the underlying model predictions, but their design and evaluation are not independent of the observed results. Bootstrap intervals and paired tests do not account for this policy-selection process.

These analyses quantify behavior on LOGIC rather than establish deployment-ready policies. Confirmatory evaluation requires fixing the evidence rules, policy definitions, thresholds, and planned comparisons before testing on independently constructed cases. The proposed distinction among supported, contradicted, and unresolved evidence must likewise be implemented and evaluated prospectively before it can support autonomous engineering decisions. Engineering use additionally requires representative revisions and review of unresolved or conflicting requests.

\section{Conclusion}
\label{sec:conclusion}

We presented LOGIC, a controlled 168-scenario benchmark for evaluating LLM intent grounding over genuine ECAD revision changes, followed by deterministic typed impact propagation. Propagating every detected revision difference without grounding is clearly insufficient. Across 504 frozen predictions from three local models, grounding capability otherwise depends on the available evidence. Gate-only achieves exact selection on explicitly anchored general cases, with token lexical also performing strongly, while the LLMs exceed both deterministic methods on the 12 relational-paraphrase cases. Their performance deteriorates as candidate inventories grow, whereas deterministic graph replay maintains comparatively stable accuracy for fixed selections over the tested graph sizes.

Evidence-based filtering suppresses false positives but can remove correct semantic selections. Exploratory evidence-empty abstention reduces unsafe reporting at the cost of positive-case coverage, with conflicting requests remaining unresolved. Overall, the results support a human-supervised hybrid architecture rather than unrestricted autonomous deployment. Future work should validate the benchmark with independent, expert-reviewed engineering cases and evaluate policies that distinguish unsupported selections from explicit contradictions.


\acknowledgements 
During the preparation of this work, the authors used OpenAI ChatGPT and Anthropic Claude to assist in various sections with language drafting, editorial refinement, and analysis support. The authors reviewed and edited all resulting material and take full responsibility for the content of this article.


\bibliographystyle{IEEEtran}
\bibliography{MyBibFile}

\thebiography
\begin{biographywithpic}
{Muhammad Faraz Shoaib}{Faraz.jpeg}
received his B.E. degree in Electrical Engineering from NEDUET, Karachi, Pakistan. He is currently a graduate student in Computer Science at Wichita State University and a Graduate Researcher at the National Institute for Aviation Research (NIAR). His research interests include artificial intelligence, large language models, agentic AI, cybersecurity, software reliability, machine learning, robotics, and intelligent engineering systems. His work explores trustworthy and reliable AI, LLM-based software analysis, security of tool-using AI agents, and the application of AI and computational methods to real-world software and engineering problems.
\end{biographywithpic}

\begin{biographywithpic}
{Muhammad Qasim}{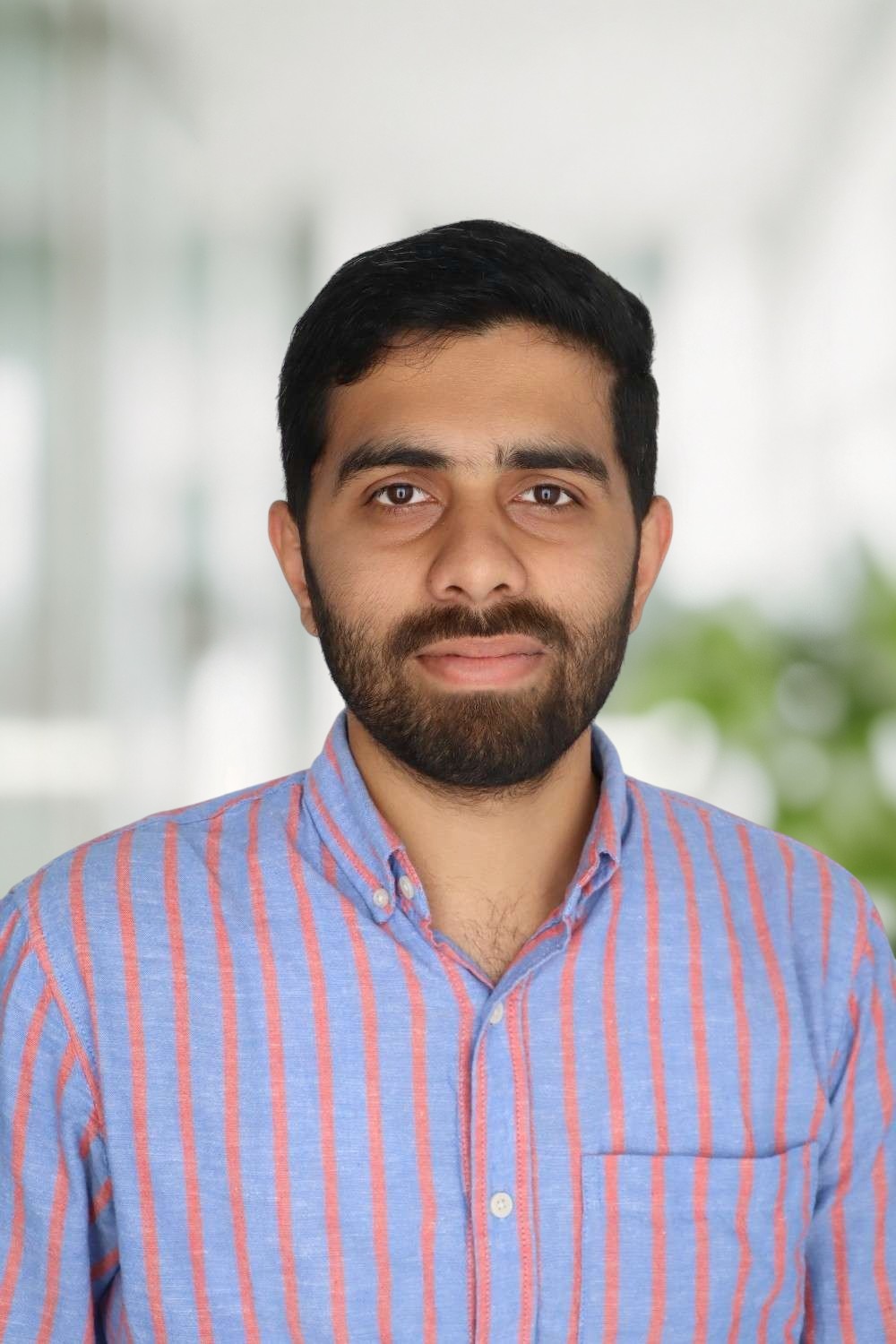}
received his B.S. degree in Computer Science from the University of Agriculture Faisalabad, Pakistan. He has more than four years of professional software engineering experience, including work in CAD/CAM-based aerospace engineering, digital product development, and engineering design workflows. He is currently a graduate student in Computer Science at Wichita State University, Wichita, Kansas, USA, and a Graduate Research Assistant at the National Institute for Aviation Research (NIAR). His research interests include large language models, agentic AI, cybersecurity, and trustworthy AI, with a focus on the reliability and security of LLM- and agent-based systems.
\end{biographywithpic}

\begin{biographywithpic}
{Raisulhaq Mohammed Rizwan}{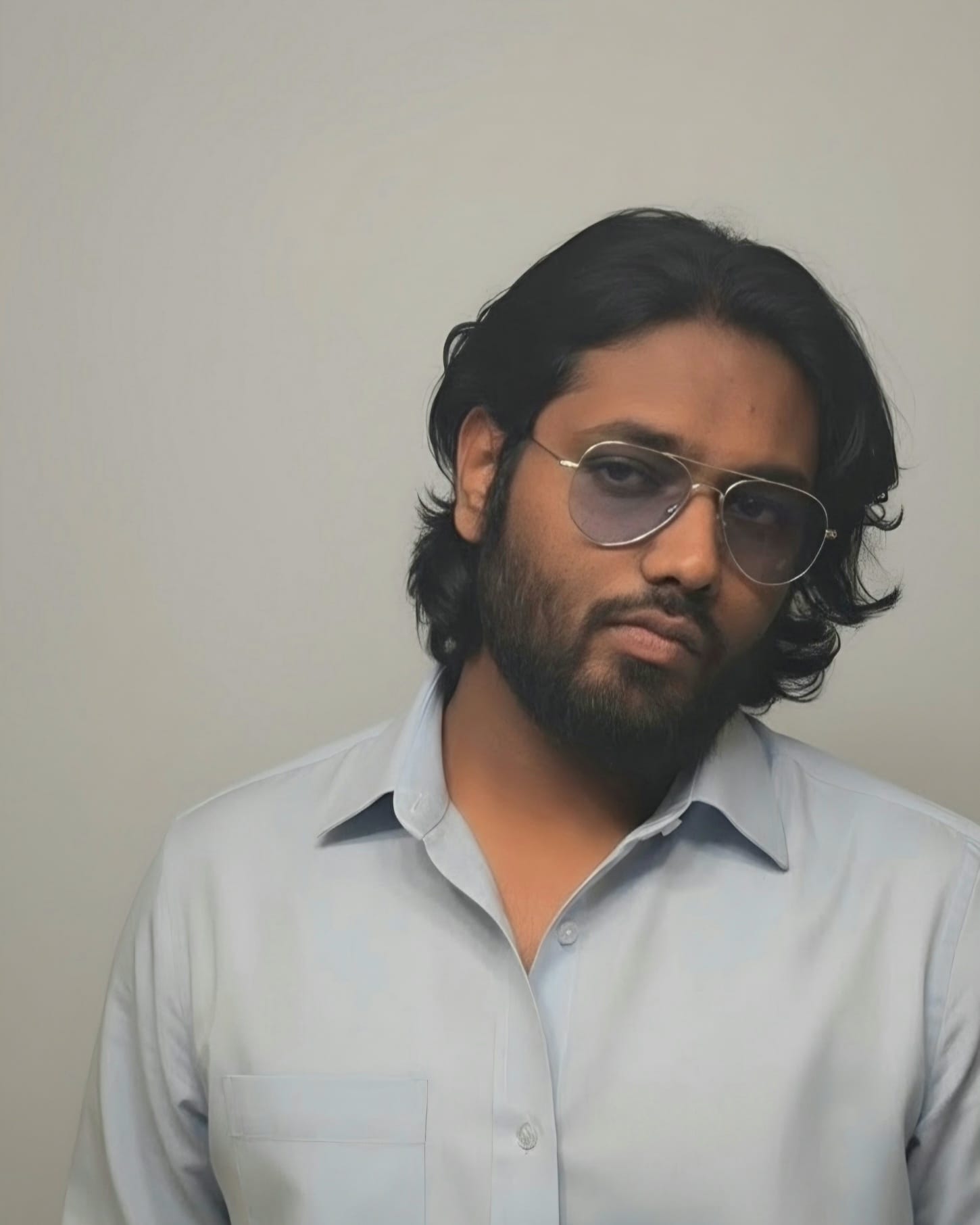}
is currently a Research Associate and Team Lead in the Digital Twin Department at the National Institute for Aviation Research, Wichita State University. He received his M.S. degree in Computing from Wichita State University and his B.E. degree in Electrical and Instrumentation Engineering from Osmania University, Hyderabad, India. His research interests include artificial intelligence, digital twins, robotics, and model-based systems engineering. His work focuses on applying large language models and graph-based methods to aerospace electrical design, configuration management, and verification. He has developed embedded robotics and unmanned aerial systems, and holds a patent for the MIMOSA SUIT safety device. 

\end{biographywithpic}

\begin{biographywithpic}
{Rahmatullah Safdar}{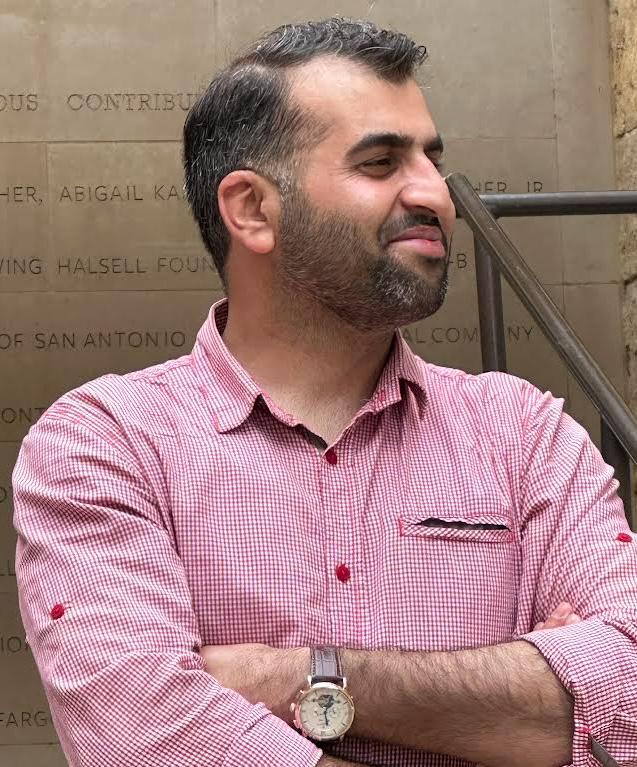}
earned his bachelor’s degree in electrical engineering from Bahria University, Islamabad, Pakistan. His interest in academia led him to pursue a graduate program at the National University of Sciences and Technology (NUST) in Islamabad, Pakistan. During his master’s in electrical engineering at NUST, he focused on developing an Artificial Pancreas Device for diabetes management. After earning his master’s in electrical engineering, Rahmat returned to his alma mater, Bahria University, as a lecturer in the Department of Electrical Engineering, where he served for one year before being accepted into the Biomedical Engineering Ph.D. program at Wichita State University. He is currently conducting research under the guidance of Dr. Nils Hakansson, focusing on the musculoskeletal biomechanics of human rolling, and leading a graduate student’s group to promote biomedical research at Wichita State and Kansas.

\end{biographywithpic}

\begin{biographywithpic}
{Abdul Aleem Mohammed}{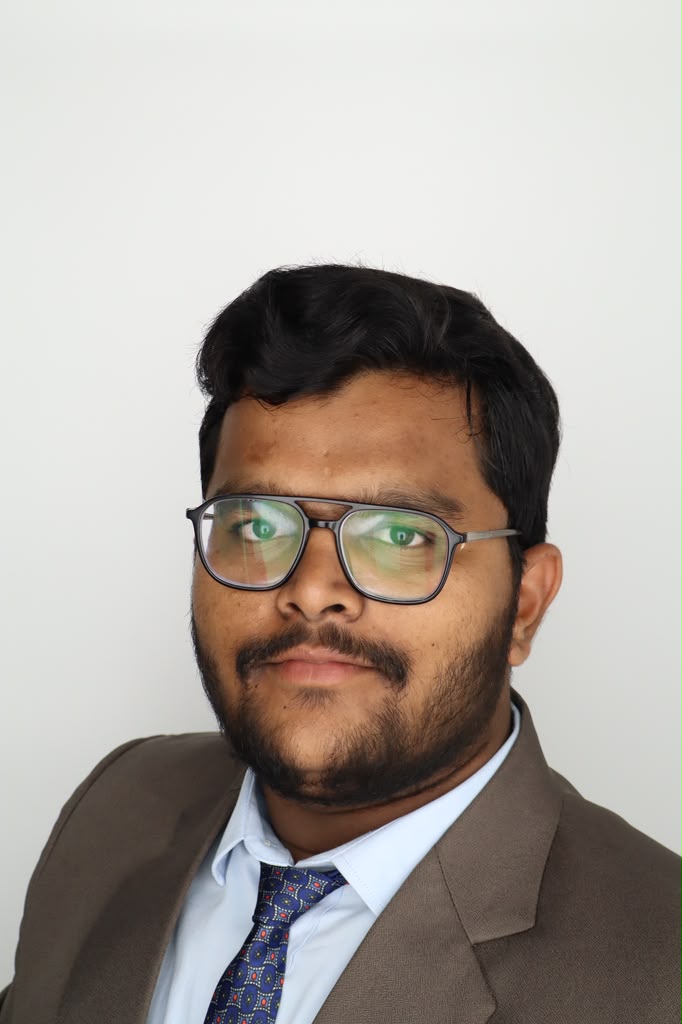}
is currently an M.S. student in Computer Science at Wichita State University. He received his B.E. degree in Computer Science from Osmania University, Hyderabad, India. His research interests include artificial intelligence, machine learning, generative AI, computer vision, and AI foundations. His research focuses on developing intelligent systems for applications in agriculture, healthcare, media authenticity, and engineering. He has contributed to research involving deep learning, intelligent IoT systems, optimization techniques, and data-driven AI solutions. He is currently involved in AI research and engineering activities at the National Institute for Aviation Research at Wichita State University.

\end{biographywithpic}

\begin{biographywithpic}
{Muzammil Adnan Shaik}{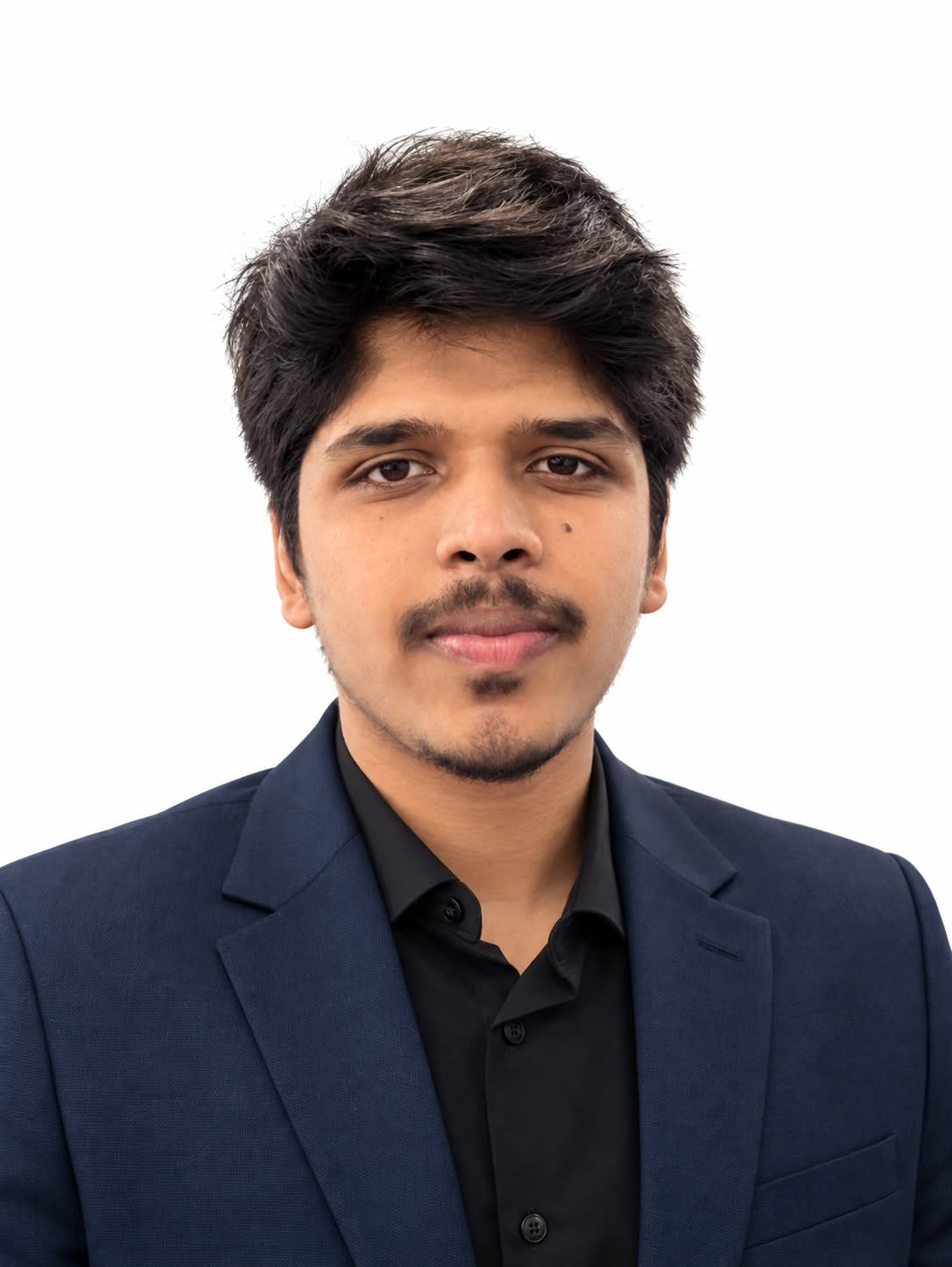}
received his B.E. degree in Mechanical Engineering and is currently an M.S. student in Industrial Engineering at Wichita State University, Wichita, Kansas, USA. He is also associated with the National Institute for Aviation Research (NIAR) at Wichita State University. His research interests include advanced manufacturing, quality and reliability engineering, industrial automation, process optimization, robotics, and data-driven engineering. His academic and professional experience spans mechanical design, manufacturing, quality engineering, robotics, automation, and data analytics. His work focuses on applying modern engineering methods and emerging technologies to improve the efficiency, quality, reliability, and performance of manufacturing and industrial systems.

\end{biographywithpic}

\end{document}